\documentclass[letterpaper, 10 pt, conference]{ieeeconf}  % Comment this line out if you need a4paper

\IEEEoverridecommandlockouts                              % This command is only needed if 
\usepackage{graphics} % for pdf, bitmapped graphics files
\usepackage{epsfig} % for postscript graphics files
\usepackage{amsmath} % assumes amsmath package installed
\usepackage{amssymb}  % assumes amsmath package installed
\usepackage{booktabs}
\usepackage{xcolor}

\usepackage{algorithmic}
\usepackage{algorithm}

\newcommand{\sO}{\mathcal{O}}
\newcommand{\sS}{\mathcal{S}}
\newcommand{\sA}{\mathcal{A}}
\newcommand{\sH}{\mathcal{H}}
\newcommand{\sZ}{\mathcal{Z}}
\newcommand{\sD}{\mathcal{D}}
\newcommand{\sM}{\mathcal{M}}

\newcommand{\sL}{\mathcal{L}}
\newcommand{\E}{\mathbb{E}}
\newcommand{\R}{\mathbb{R}}

\newcommand{\algorithmicinput}{\textbf{Input:}}
\newcommand{\INPUT}{\item[\algorithmicinput]}
\newcommand{\algorithmicoutput}{\textbf{Output:}}
\newcommand{\OUTPUT}{\item[\algorithmicoutput]}

\title{\LARGE \bf
BIFROST: Bridging Invariant Feature Representation for Observation-space Sim2Real Transfer
}

\author{Yunfu~Deng,~and~Josiah~P.~Hanna
\thanks{Yunfu Deng and Josiah P. Hanna are with the Department of Computer Sciences, University of Wisconsin--Madison, Madison, WI 53706 USA (Email: yunfu.deng@wisc.edu, jphanna@cs.wisc.edu).}}%

\begin{document}

\maketitle
\thispagestyle{empty}
\pagestyle{empty}

%%%%%%%%%%%%%%%%%%%%%%%%%%%%%%%%%%%%%%%%%%%%%%%%%%%%%%%%%%%%%%%%%%%%%%%%%%%%%%%%g
\begin{abstract}
% Sim2real transfer for robot policy learning suffers from reality gaps including perceptual and dynamics gap.
% % % 
% Despite different types of reality gaps, the basis for attempting sim2real in the first place is that there is shared structure between a task in simulation and reality.
% % %
% In this paper, we study whether we can identify and exploit shared structure to train a policy that enables zero-shot transfer.
Sim2real transfer for robot policy learning suffers due to mismatch between simulation and reality. Existing methods typically address each gap in isolation through separate adaptation modules, which are composed or layered when both gaps coexist. Yet the basis for attempting sim2real in the first place is that there is shared structure between a task in simulation and reality, where equivalent actions from equivalent configurations produce equivalent long term outcomes regardless of domain specific differences in rendering or physics. In this paper, we study whether we can identify and exploit this shared structure from raw observations to train a policy that enables zero shot transfer.
We introduce BIFROST, which learns a shared history encoder on paired cross-domain data via cross-domain bisimulation objective: observation-action sequences leading to equivalent long-term behavior are mapped to nearby latent states, regardless of domain. 
Policies trained on these latent states in simulation transfer zero-shot to reality. We provide empirical evidence on sim2sim visual navigation and sim2real contact rich manipulation task and visual servoing task that BIFROST achieves effective transfer where domain adaptation and co-training baselines fail under both visual and dynamics domain gaps.
\end{abstract}

%%%%%%%%%%%%%%%%%%%%%%%%%%%%%%%%%%%%%%%%%%%%%%%%%%%%%%%%%%%%%%%%%%%%%%%%%%%%%%%%


%%%%%%%%%%%%%%%%%%%%%%%%%%%%%%%%%%%%%%%%%%%%%%%%%%%%%%%%%%%%%%%%%%%%%%%%%%%%%%%%

%%%%%%%%%%%%%%%%%%%%%%%%%%%%%%%%%%%%%%%%%%%%%%%%%%%%%%%%%%%%%%%%%%%%%%%%%%%%%%%%

% \input{chapters/introduction}

\section{Introduction}
Modern robot learning increasingly relies on large-scale training in simulation~\cite{tobin2017domain, peng2018sim, rudin2022learning}, as real-world data remains expensive, slow, or operationally risky to collect. The practical utility of sim2real transfer, however, hinges on how faithfully the simulator reproduces the aspects of the real world that matter for control. In practice, simulators inevitably diverge from reality in two respects: rendered images differ from camera images (the perceptual gap), and physics engines approximate contact, friction, and actuation imperfectly (the dynamics gap)~\cite{zhao2020sim}. These compounded discrepancies remain a central obstacle in deploying simulation-trained policies on physical hardware. A large body of work addresses each gap source in isolation: system identification and dynamics randomization target the dynamics gap~\cite{yu2017preparing, chebotar2019closing, peng2018sim}, while visual randomization and domain adaptation target the perceptual gap~\cite{tobin2017domain, ho2021retinagan}. When both gaps coexist, however, these methods must be composed or layered~\cite{truong2021bi}, and the interactions between gap sources are left unaddressed.

Sim2real transfer rests on an implicit assumption: the source and target domains, despite differing in observations and dynamics, share a common low-dimensional latent structure. Analogous assumptions underpin multitask representation learning in reinforcement learning, in which multiple MDPs that share a low-rank transition structure admit a common representation under which a single policy generalizes across tasks~\cite{agarwal2020flambe, cheng2022provable}. Viewing simulation and reality as two such MDPs, the problem reduces to learning this shared representation from cross-domain data. This reduction motivates the central question of this work:
\begin{center}
\textit{``How can we learn a shared representation from raw observations that enables zero-shot policy transfer across domains?''}
\end{center}

A natural answer is to define cross-domain similarity through behavioral equivalence: two states from different domains are similar when acting identically from both leads to equivalent long-term outcomes. The generalized bisimulation metric (GBSM)~\cite{tao2025a} provides a principled foundation for measuring such cross-domain state similarity: it quantifies closeness through reward equivalence and transition equivalence between pairs of MDPs, guaranteeing that states closer under this metric have more similar value functions. GBSM, however, requires access to true states in both domains. Methods such as DBC~\cite{zhang2020learning} and MICo~\cite{castro2021mico} have converted bisimulation into a representation learning objective that operates from observations, but are restricted to a single MDP. The key insight of this work is that cross-domain bisimulation alignment, carried out in a learned latent space from observation histories, can unify perception and dynamics gaps into a single representation learning problem.

Building on this insight, we propose BIFROST, Bridging Invariant Feature Representation for Observation-space Sim2Real Transfer. BIFROST learns a shared history encoder that maps observation-action sequences from both domains into a common latent space via a cross-domain bisimulation objective, and trains a control policy on this latent space for zero-shot sim2real transfer. The contributions of our work are two-fold. First, we introduce cross-domain bisimulation alignment as a representation learning mechanism for sim2real transfer that addresses both perception and dynamics gaps, without requiring access to true states in either domain. Second, we validate the framework on both navigation and contact-rich manipulation tasks, including sim2sim experiments with controlled gap sources and sim2real transfer on physical hardware, demonstrating effective zero-shot transfer in settings in which existing domain adaptation and co-training baselines fail under compounded domain gaps.

\vspace{-1mm}
\section{Related Work}
This section reviews sim2real transfer methods that address the domain gap through simulation engineering, and theoretical foundations of behavioral equivalence and state abstraction that underpin our approach.

\subsection{Sim2Real via Calibration, Randomization, and Distillation}

A large body of sim2real work can be organized around three complementary strategies for addressing the reality gap: (i) reducing the simulator-reality mismatch through system identification, (ii) tolerating mismatch by training policies that are robust across a family of environments, and (iii) exploiting simulator-only privileged information to train policies deployable under real-world sensing constraints.

The first strategy, system identification, aims to make the simulator as close to reality as possible. Yu et al.~\cite{yu2017preparing} calibrate explicit physical parameters, while Chebotar et al.~\cite{chebotar2019closing} use neural networks to model complex dynamics and actuator behavior. Truong et al.~\cite{truong2021bi} learn separate visual and dynamics adaptors, which can also be viewed through this lens. Ho et al.~\cite{ho2021retinagan} instead reduce the perceptual mismatch by translating simulated images toward photorealism with RetinaGAN. 
% More recently, Byravan et al.~\cite{byravan2022nerf2real} and Qureshi et al.~\cite{qureshi2025splatsim} reconstruct real scenes photorealistically using neural radiance fields and Gaussian Splatting, respectively, further narrowing the visual gap.

The second strategy, domain randomization (DR), introduced by Tobin et al.~\cite{tobin2017domain}, trains policies to be robust to the gap rather than eliminating it. DR is applied to dynamics parameters~\cite{peng2018sim, hwangbo2019learning}, visual rendering~\cite{sadeghi2016cad2rl}, or both simultaneously~\cite{andrychowicz2020learning, li2025reinforcement}.

The third strategy exploits the asymmetry between simulation and deployment: simulators provide access to privileged information, including ground-truth state, that is unavailable in the real world. Pinto et al.~\cite{pinto2017asymmetric} propose an asymmetric actor-critic that grants the critic full-state access while restricting the actor to partial observations. Teacher-student frameworks~\cite{lee2020learning, rudin2022learning, miki2022learning} extend this idea by distilling a state-privileged teacher into a student policy that operates from partial observations.

These three strategies intervene by modifying the simulator, the training distribution, or the learning signal, respectively. A complementary perspective studies transfer through behavioral equivalence, formalizing when domain differences are irrelevant for control.

\subsection{Behavioral Equivalence and State Abstraction}

Abstraction-based approaches aim to identify control-relevant invariances that are stable across domain differences, thereby supporting transfer without requiring the simulator to be fully faithful. The theoretical foundation originates with probabilistic bisimulation, in which two states are deemed equivalent if they yield identical reward distributions and transition to equivalent successor states. Ferns et al.~\cite{ferns2004metrics, ferns2011bisimulation} extended this binary notion to continuous bisimulation metrics, establishing that states close under the metric have similar optimal value functions. Subramanian et al.~\cite{subramanian2022approximate} developed the framework of Approximate Information States, showing that representations satisfying approximate information-state conditions admit bounded optimality loss. The algebraic perspective of MDP homomorphisms~\cite{ravindran2003smdp, rezaei2022continuous} provides a complementary view: states with zero bisimulation distance form equivalence classes that define a reduced quotient MDP, and policies can be lifted between the original and abstract models.

These formulations, however, operate within a single MDP and do not address cross-domain comparison. Castro and Precup~\cite{castro2010using} leveraged lax bisimulation metrics to measure distances between states in distinct MDPs, enabling policy transfer by identifying behaviorally similar state-action pairs across domains. Fickinger et al.~\cite{fickinger2021cross} use the Gromov-Wasserstein distance for cross-domain imitation but require well-defined metrics on low-dimensional state spaces in both domains. Tao et al.~\cite{tao2025a} recently formalized the Generalized Bisimulation Metric (GBSM) between arbitrary MDP pairs, proving an inter-MDP triangle inequality and deriving tighter transfer bounds than those available from single-MDP metrics. These cross-domain formulations substantially expand the applicability of bisimulation to transfer settings, yet they continue to assume direct access to the underlying state.

Complementary to the bisimulation perspective, a growing body of work provides theoretical~\cite{cheng2022provable, brunskill2013sample, agarwal2020flambe} and empirical~\cite{maddukuri2025simreal, wei2025empirical} support for co-training shared representations across simulation and reality, showing that abundant simulation data can reduce real-world sample requirements when the two domains share underlying structure. BIFROST builds on these two lines: it extends bisimulation-based behavioral equivalence from the state space to a learned latent space, and leverages co-training on paired cross-domain data to accommodate partial observability.

\section{Method}
In this section, we present BIFROST, a framework for zero-shot sim2real transfer from raw observations. BIFROST addresses sim2real transfer through a two-phase pipeline. In the first phase, a shared history encoder is trained on paired offline data from both domains, using a cross-domain bisimulation objective that enforces reward equivalence and transition alignment in a learned latent space. In the second phase, the encoder is frozen, and a control policy is trained via online reinforcement learning in simulation with encoded histories as inputs. At deployment, target-domain observations pass through the same frozen encoder, producing latent states from which the policy acts without further adaptation.

\subsection{Problem Formulation}

We formalize the sim2real transfer problem under the framework of a Partially Observable Markov Decision Process (POMDP), defined by the tuple $\langle \sS, \sA, P, r, \sO, f, \gamma \rangle$, where $\sS$ is the state space, $\sA$ is the action space, $P: \sS \times \sA \to \Delta(\sS)$ specifies the transition dynamics, $r: \sS \times \sA \to \mathbb{R}$ is the reward function, $\sO$ is the observation space, $f: \sS \to \sO$ is a deterministic sensing function, and $\gamma \in [0,1)$ is the discount factor. Because $f$ is not invertible, multiple distinct states may yield an identical observation, and the agent cannot recover the full state from a single observation.

In visual robotic tasks, the observation decomposes into two modalities: $o_i = (o_{v,i},\, o_{p,i})$, where $o_{v,i} \in \mathbb{R}^{H \times W \times C}$ denotes visual input such as RGB or depth images and $o_{p,i} \in \mathbb{R}^{d_p}$ denotes proprioceptive readings such as joint angles or end-effector poses. Note that $o_{p,i} \neq s_i$: the true state $s_i$ encompasses unobservable quantities such as friction coefficients and contact forces, which necessitates the integration of visual information to infer the latent dynamics. The agent seeks a policy $\pi$ that maximizes the expected discounted return:
\begin{equation}
\label{eq:objective}
J_{\sM}(\pi) \;=\; \mathbb{E}\!\left[\sum_{i=0}^{\infty} \gamma^{i}\, r(s_i, a_i)\right].
\end{equation}

The sim2real setting involves two POMDPs, a source domain (simulation) $\sM_{\mathrm{src}}$ and a target domain (real world) $\sM_{\mathrm{tgt}}$, that share the same $\sA$ and $\gamma$ but differ in two respects. First, the dynamics gap: the transition functions differ, $P_{\mathrm{src}} \neq P_{\mathrm{tgt}}$, due to approximations in the physics engine such as simplified friction or unmodeled delays. Second, the perceptual gap: for the same underlying state $s$, the visual observations differ, $o_{v,\mathrm{src}} \neq o_{v,\mathrm{tgt}}$, owing to discrepancies in textures, lighting, and rendering. Because $f$ is not invertible and the true state includes quantities unobservable from a single timestep, the agent must condition on the observation-action history $h_t = (o_0, a_0, \ldots, o_t)$ rather than a single observation.
A history encoder compresses $h_t$ into a low-dimensional latent state $z_t$ from which the policy acts. The encoder design is detailed in Section~\ref{sec:representation}.

\begin{figure*}[!t]
    \centering
    \includegraphics[width=0.97\textwidth]{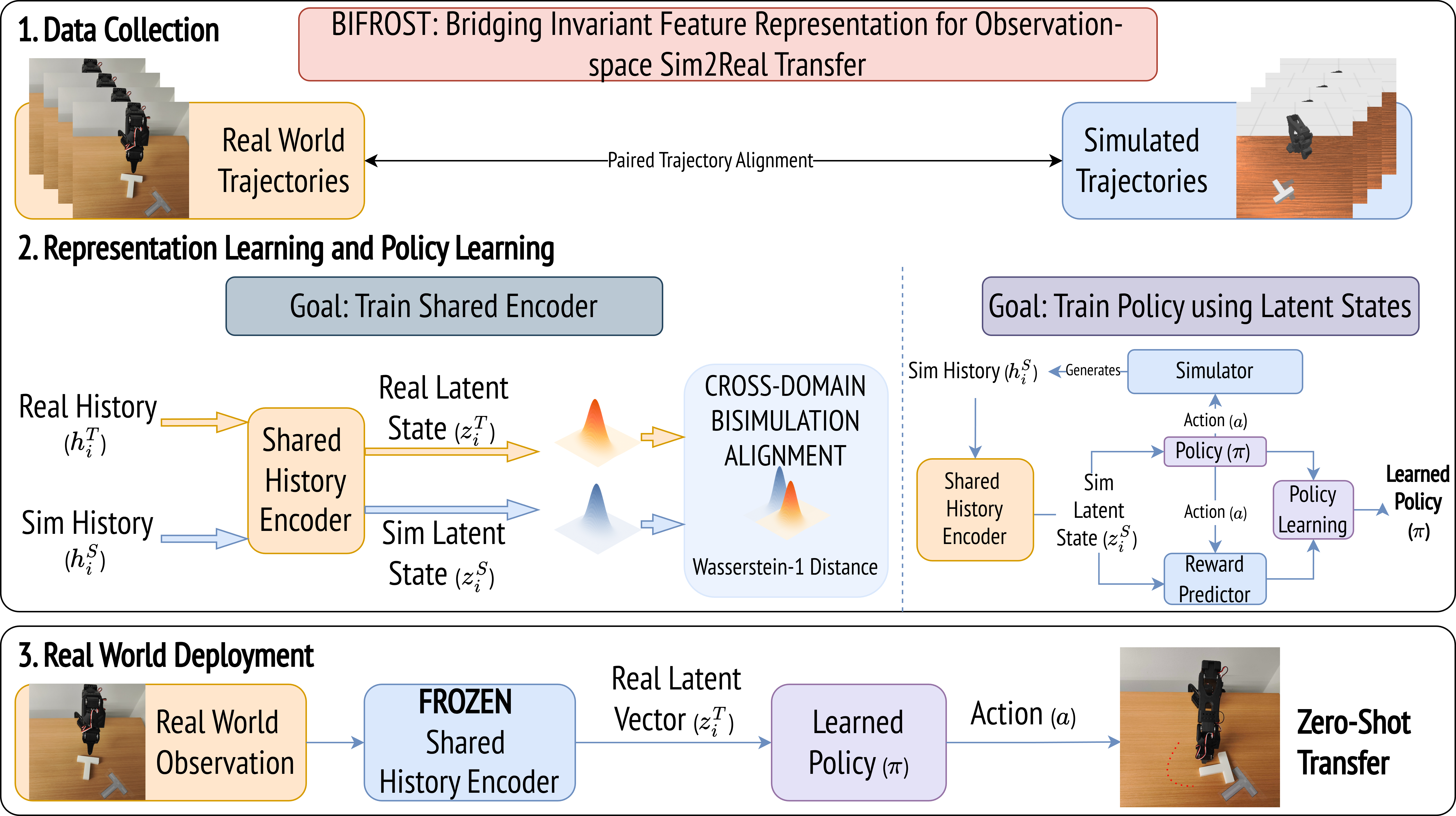}
    \caption{Overview of the BIFROST framework. A shared history encoder is trained on paired cross-domain data so that observation-action sequences leading to equivalent long-term behavior are mapped to nearby latent states, regardless of domain-specific visual or dynamics differences. The policy is trained and deployed in the same latent space, enabling zero-shot transfer without separate treatment of individual gap sources.}
    \label{fig:pipeline}
\end{figure*}

\subsection{Cross-Domain Latent Alignment}

BIFROST learns a shared history encoder $\phi$ that maps observation-action sequences from both domains into a common latent space, such that histories yielding similar rewards and similar latent successor distributions are mapped to nearby latent states regardless of their domain of origin. The encoder is trained on paired offline data from simulation and reality with three objectives: a reward prediction loss, a latent dynamics prediction loss, and a cross-domain alignment loss. This explicit alignment is necessary because co-training alone, i.e., training a shared encoder on pooled data from both domains, does not guarantee that behaviorally equivalent cross-domain histories are mapped nearby, particularly when visual and dynamics discrepancies are substantial.

The design of these objectives is motivated by the Generalized Bisimulation Metric (GBSM) between pairs of MDPs~\cite{tao2025a}. For MDPs sharing the same action space and discount factor, GBSM defines cross-domain state distance as $d(s_1,s_2) = \max_a \{ |R_{\mathrm{src}}(s_1,a) - R_{\mathrm{tgt}}(s_2,a)| + \gamma\, W_1(P_{\mathrm{src}}(\cdot|s_1,a),\, P_{\mathrm{tgt}}(\cdot|s_2,a);\, d) \}$, where $s_1 \in \sS_{\mathrm{src}}$, $s_2 \in \sS_{\mathrm{tgt}}$, the metric is recursive with $d$ serving as its own ground cost, and satisfies $|V^*_{\mathrm{src}}(s_1) - V^*_{\mathrm{tgt}}(s_2)| \leq d(s_1,s_2)$.

Our three losses target the two components of this metric in the learned latent space: $\sL_{\mathrm{reward}}$ reduces the reward discrepancy between paired latent states, $\sL_{\mathrm{align}}$ minimizes the Wasserstein distance between their predicted successor distributions, and $\sL_{\mathrm{LDP}}$ trains the forward dynamics model that $\sL_{\mathrm{align}}$ relies on. 
Our formulation approximates rather than exactly computes GBSM: GBSM is defined over MDP states, whereas our setting is a POMDP in which the true state is not accessible. Following the Approximate Information State framework~\cite{subramanian2022approximate}, we replace states encodings with history encodings that satisfy approximate reward prediction and approximate dynamics prediction conditions. Nevertheless, the value function bound implies that latent states close under the learned metric share similar optimal value functions, supporting zero-shot policy transfer.

\subsection{Paired Trajectory Collection}
\label{sec:paired}

Learning cross-domain representations requires correspondence between observations from simulation and reality. We first collect a dataset $\sD_{\mathrm{tgt}}$ in the target domain. For each trajectory $\tau \in \sD_{\mathrm{tgt}}$, $n_\tau$ segments of length $l$ are sampled at uniformly random starting indices. 
The corresponding source-domain segment is generated by initializing the simulator state with the observable configuration $\xi_i$, such as joint angles and object poses, extracted from $o^{\mathrm{tgt}}_i$, while unobservable state variables, such as global position under egocentric observation and contact forces, evolve from the simulator's forward dynamics. Environment parameters such as friction coefficients and actuator delays, which define the MDP rather than its state, remain at default values $\theta_{\mathrm{def}}$.
The same action sequence $\{a_i, \ldots, a_{i+l-1}\}$ recorded from the target trajectory is then replayed in the simulator. This procedure, summarized in Algorithm~\ref{alg:paired}, yields a paired dataset $\sD_{\mathrm{paired}} = \{(o^{\mathrm{src}}_k, o^{\mathrm{tgt}}_k, a_k, r^{\mathrm{tgt}}_k)\}$, where $k$ indexes corresponding timesteps across domains. Because both segments share the same initial configuration and action sequence, subsequent divergence in observations can be attributed to domain-specific discrepancies in rendering and physics rather than differences in task-relevant state.

\begin{algorithm}[t]
\caption{Paired Trajectory Collection}
\label{alg:paired}
\begin{algorithmic}[1]
\INPUT Target dataset $\sD_{\mathrm{tgt}}$, simulator $\sM_{\mathrm{src}}$, segment length $l$, default parameters $\theta_{\mathrm{def}}$
\OUTPUT Paired dataset $\sD_{\mathrm{paired}}$
\STATE $\sD_{\mathrm{paired}} \leftarrow \emptyset$
\FOR{each trajectory $\tau \in \sD_{\mathrm{tgt}}$}
    \STATE Sample $n_{\tau}$ starting indices $\{i_1, \ldots, i_{n_\tau}\}$ uniformly from $\{0, \ldots, |\tau| - l\}$
    \FOR{each starting index $i$}
        \STATE Extract observable configuration $\xi_i$ from $o^{\mathrm{tgt}}_i$
        \STATE Initialize $\sM_{\mathrm{src}}$ at $(\xi_i,\, \theta_{\mathrm{def}})$
        \STATE $\tau_{\mathrm{seg}} \leftarrow \emptyset$
        \FOR{$k = i$ \TO $i + l - 1$}
            \STATE Execute $a_k$ in $\sM_{\mathrm{src}}$, observe $o^{\mathrm{src}}_k$
            \STATE $\tau_{\mathrm{seg}} \leftarrow \tau_{\mathrm{seg}} \cup \{(o^{\mathrm{src}}_k,\, o^{\mathrm{tgt}}_k,\, a_k,\, r^{\mathrm{tgt}}_k)\}$
        \ENDFOR
        \STATE $\sD_{\mathrm{paired}} \leftarrow \sD_{\mathrm{paired}} \cup \{\tau_{\mathrm{seg}}\}$
    \ENDFOR
\ENDFOR
\end{algorithmic}
\end{algorithm}

\subsection{Cross-Domain Representation Learning}
\label{sec:representation}

Given the paired dataset $\sD_{\mathrm{paired}}$, the shared history encoder $\phi: \sH \to \sZ$ is trained to map observation-action sequences from both domains into a common latent space. The encoder is implemented as a GRU network that consumes the history $h_t = (o_0, a_0, \ldots, o_t)$ and outputs a latent state $z_t = \phi(h_t)$. The three loss components are defined as follows; all expectations are taken over paired segments sampled from $\sD_{\mathrm{paired}}$.

For the reward component, a predictor $\widehat{R}: \sZ \times \sA \to \R$ is trained to estimate the target-domain reward from latent states of both domains:
\begin{equation}
    \sL_{\mathrm{reward}} = \E \bigl[ (\widehat{R}(\phi(h^{\mathrm{src}}), a) - r^{\mathrm{tgt}})^2 + (\widehat{R}(\phi(h^{\mathrm{tgt}}), a) - r^{\mathrm{tgt}})^2 \bigr].
\end{equation}
Supervising both source and target representations with $r^{\mathrm{tgt}}$ ensures that the encoder prioritizes features predictive of real-world task outcomes.

For the transition component, a latent forward model $\widehat{P}: \sZ \times \sA \to \Delta(\sZ)$ is parameterized as a conditional Gaussian: $\widehat{P}(\cdot \mid z, a) = \mathcal{N}(\mu(z,a),\, \sigma^2(z,a)I)$. Two objectives are imposed on this model. A latent dynamics prediction (LDP) loss trains the forward model to predict the next latent state within each domain, while simultaneously shaping the encoder so that successive latent states are dynamically coherent:
\begin{equation}
    \sL_{\mathrm{LDP}} = \E \left[ \frac{\|\mu(z_i, a_i) - z_{i+1}\|^2}{2\sigma^2(z_i, a_i)} + \frac{d_z}{2} \ln \sigma^2(z_i, a_i) \right],
\end{equation}
where $d_z$ is the latent dimension. An alignment loss minimizes the Wasserstein-1 distance between the latent successor distributions of paired source and target histories:
\begin{equation}
    \sL_{\mathrm{align}} = \E \left[ W_1\!\left( \widehat{P}(\cdot \mid \phi(h^{\mathrm{src}}), a),\; \widehat{P}(\cdot \mid \phi(h^{\mathrm{tgt}}), a) \right) \right].
\end{equation}
In practice, the $W_1$ distance is approximated via the Kantorovich-Rubinstein duality~\cite{villani2008optimal} using a learned 1-Lipschitz critic $f_\psi$:
\begin{equation}
\begin{aligned}
    W_1\!\left(\widehat{P}(\cdot \mid z,a),\; \widehat{P}(\cdot \mid z',a)\right)
    &\approx \sup_{\|f_\psi\|_L \le 1}
    \E_{z^+ \sim \widehat{P}}\!\left[ f_\psi(z^+) \right] \\
    &\quad - \E_{z'^+ \sim \widehat{P}}\!\left[ f_\psi(z'^+) \right],
\end{aligned}
\end{equation}
where $z^+, z'^+$ are sampled via the reparameterization trick and the Lipschitz constraint is enforced by penalizing $(\|\nabla_{z} f_\psi(z)\|_2 - 1)^2$ on interpolated samples between $z^+$ and $z'^+$.
The total representation learning objective is
\begin{equation}
    \sL = \sL_{\mathrm{reward}} + \lambda_{\mathrm{LDP}}\,\sL_{\mathrm{LDP}} + \lambda_{\mathrm{align}}\,\sL_{\mathrm{align}},
\end{equation}
where $\lambda_{\mathrm{LDP}}$ and $\lambda_{\mathrm{align}}$ are weighting coefficients.

\subsection{Policy Learning and Deployment}
\label{sec:policy}

Once the encoder $\phi$ is trained, its parameters are frozen. A control policy $\pi: \sZ \to \sA$ is then trained via online reinforcement learning in the source domain $\sM_{\mathrm{src}}$. At each timestep, the history $h_t$ is encoded into a latent state $z_t = \phi(h_t)$, and the policy selects actions based solely on $z_t$. The learned reward predictor $\widehat{R}(z_t, a_t)$ from Section~\ref{sec:representation} provides the reward signal. Because the policy operates entirely in the latent space, both in its state representation and its reward, it is never exposed to domain-specific visual or dynamics features.

The cross-domain alignment produces a shared latent space in which behaviorally equivalent histories from both domains occupy nearby regions. A policy trained on source latent states therefore generalizes to target latent states. At deployment in $\sM_{\mathrm{tgt}}$, the same frozen encoder maps target-domain histories into this latent space, enabling zero-shot transfer without further adaptation.

\begin{algorithm}[t]
\caption{BIFROST Training and Deployment}
\label{alg:bifrost}
\begin{algorithmic}[1]
\INPUT Paired dataset $\sD_{\mathrm{paired}}$ (Alg.~\ref{alg:paired}), source environment $\sM_{\mathrm{src}}$, critic steps $n_{\mathrm{critic}}$, weights $\lambda_{\mathrm{LDP}},\, \lambda_{\mathrm{align}}$
\OUTPUT Trained policy $\pi$ and frozen encoder $\phi$
\STATE Initialize encoder $\phi$, reward predictor $\widehat{R}$, forward model $\widehat{P}$, critic $f_\psi$
\STATE \textit{// Phase 1: Cross-Domain Representation Learning}
\FOR{each training iteration}
    \STATE Sample batch of paired segments from $\sD_{\mathrm{paired}}$
    \STATE Encode histories: $z^{\mathrm{src}} = \phi(h^{\mathrm{src}})$,\; $z^{\mathrm{tgt}} = \phi(h^{\mathrm{tgt}})$
    \FOR{$n_{\mathrm{critic}}$ steps}
        \STATE Sample $z^+ \!\sim\! \widehat{P}(\cdot \mid z^{\mathrm{src}}, a)$,\; $z'^+ \!\sim\! \widehat{P}(\cdot \mid z^{\mathrm{tgt}}, a)$
        \STATE Update $f_\psi$ to maximize $\E[f_\psi(z^+)] - \E[f_\psi(z'^+)]$ with gradient penalty on interpolations between $z^+$ and $z'^+$
    \ENDFOR
    \STATE Compute $\sL_{\mathrm{reward}}$, $\sL_{\mathrm{LDP}}$, $\sL_{\mathrm{align}}$
    \STATE Update $\phi,\, \widehat{R},\, \widehat{P}$ on $\sL_{\mathrm{reward}} + \lambda_{\mathrm{LDP}}\,\sL_{\mathrm{LDP}} + \lambda_{\mathrm{align}}\,\sL_{\mathrm{align}}$
\ENDFOR
\STATE \textit{// Phase 2: Policy Learning in Source Domain}
\STATE Freeze $\phi$ and $\widehat{R}$
\STATE Train $\pi: \sZ \to \sA$ via SAC in $\sM_{\mathrm{src}}$, using $z_t = \phi(h_t)$ as state and $\widehat{R}(z_t, a_t)$ as reward
\STATE \textit{// Deployment}
\STATE In $\sM_{\mathrm{tgt}}$: execute $\pi(\phi(h^{\mathrm{tgt}}_t))$ with the same frozen $\phi$
\end{algorithmic}
\end{algorithm}

\section{Experiments}
\label{sec:experiments}

BIFROST is evaluated on sim2sim transfer under controlled visual and dynamics gaps and sim2real transfer on physical hardware, followed by an ablation study. Our evaluation targets methods that learn transferable representations from raw observations using limited target-domain data. Simulation-only approaches such as domain randomization~\cite{tobin2017domain} and privileged-information adaptation~\cite{kumar2021rma} address a complementary setting in which no target data is used; they are orthogonal to BIFROST and can in principle be combined with it. Cross-domain optimal transport methods such as GWIL~\cite{fickinger2021cross} require ground-truth state access, which is unavailable in our partially observable setting.

Under these constraints, BIFROST is compared against five baselines: \textit{Direct Transfer}, which trains in $\sM_{\text{src}}$ and deploys zero-shot; \textit{Target-Only}, which trains via ~\cite{kostrikov2022offline} exclusively on $\sD_{\text{tgt}}$; \textit{BDA}~\cite{truong2021bi}, which learns separate visual and dynamics adaptors via adversarial training; \textit{Co-Training (BC)}~\cite{maddukuri2025simreal}, which performs behavioral cloning on expert-quality trajectories from both domains; and \textit{Co-Training (Offline RL)}, which applies IQL on the full combined dataset $\sD_{\text{src}} \cup \sD_{\text{tgt}}$. BIFROST uses the same dataset as Co-Training (Offline RL), isolating the effect of cross-domain bisimulation alignment. All methods that use target data share the same data budget. Results report zero-shot performance in $\sM_{\text{tgt}}$ as mean $\pm$ std over 10 seeds (sim2sim) and 3 seeds (sim2real).

\subsection{Sim2Sim Transfer: Visual Navigation}

Both experiments share an identical maze layout with three color-coded goals and geodesic-distance-based dense reward. The agent must navigate to a randomly assigned goal under partial observability. The target-domain data consists of 200 trajectories, yielding approximately 4,600 paired segments of average length 32.
\subsubsection{Cross-Fidelity Transfer under Top-Down Observation}

The source domain is a 2D maze rendered via OpenCV, where the agent moves via direct velocity commands. The target domain is a MuJoCo-based 3D maze where a PD-controlled point mass is observed through a tracking overhead camera. In both environments, the agent receives a $64 \times 64$ RGB crop centered on its position along with a 3-dimensional RGB vector indicating the target goal. The action space is a 2D velocity command $[v_x, v_y]$; in the source domain the agent instantaneously reaches the commanded velocity, whereas in the target domain the point mass is subject to acceleration and inertia. The dominant challenge is the extreme visual gap: the two domains share no visual primitives, with flat geometric shapes in the source and 3D surfaces with shading in the target. Successful transfer must therefore rely on learned behavioral equivalence rather than visual similarity. Since neither domain provides depth observations, BDA trains its visual adapter on RGB only.

\begin{figure}[t]
    \centering
    \includegraphics[width=0.49\linewidth]{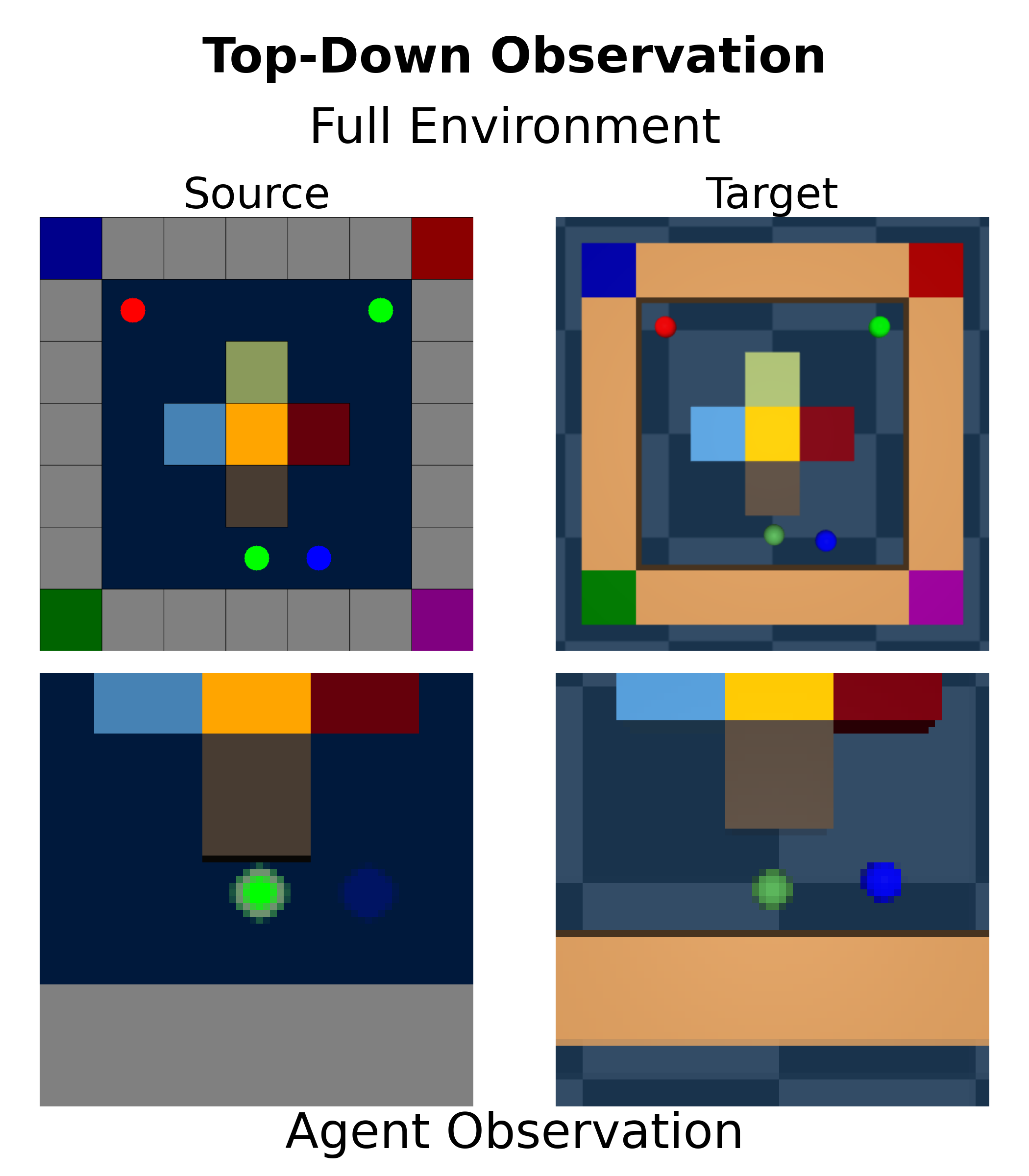}\hfill
    \includegraphics[width=0.49\linewidth]{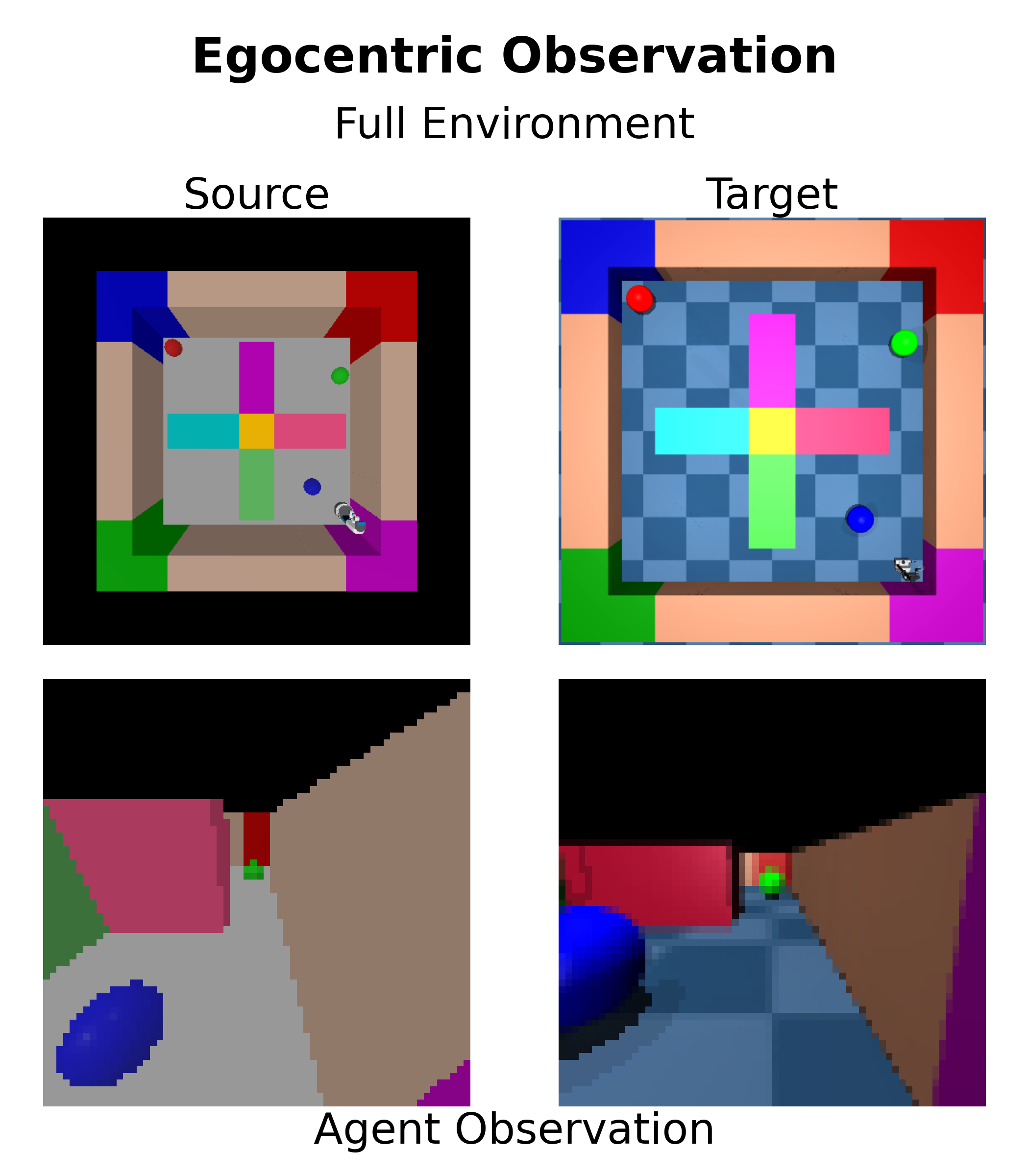}
    \caption{Left: top-down maze navigation. Right: egocentric maze navigation. In both settings, the agent observes the bottom-row $64 \times 64$ image crop as a partial observation.}
    \label{fig:env_combined}
\end{figure}

\subsubsection{Cross-Simulator Transfer under Egocentric Observation}

The source domain is a SAPIEN-based environment and the target domain is a MuJoCo environment, both using a Hello Robot Stretch navigating the same maze layout with an onboard forward-facing camera. The observation consists of a $64 \times 64$ RGB-D image, a 3-dimensional RGB vector indicating the target goal, and a 3-dimensional body-frame velocity vector obtained from proprioceptive sensors (IMU and odometry), as is standard on most mobile robot platforms. Although the robot model is shared, the two simulator backends introduce both visual and dynamics gaps: SAPIEN and MuJoCo produce visually distinct renderings of the same scene, and SAPIEN employs a simplified kinematic model for base locomotion, whereas MuJoCo simulates full rigid-body dynamics with wheel-ground contact, inertia, and friction. Compared to the top-down setting, egocentric observation further imposes stronger partial observability: the agent sees only a forward-facing view and must maintain spatial memory to navigate, directly testing the capacity of the history encoder $\phi$.

\begin{table}[t]
\centering
\caption{Sim2sim navigation results (success rate over 10 seeds).}
\label{tab:sim2sim}
\begin{tabular}{lcc}
\toprule
Method & Top-Down & Egocentric \\
\midrule
Direct Transfer          & $0.19 \pm 0.04$ & $0.03 \pm 0.02$ \\
Target-Only              & $0.46 \pm 0.09$ & $0.16 \pm 0.07$ \\
BDA                      & $0.67 \pm 0.05$ & $0.34 \pm 0.05$ \\
Co-Training (BC)         & $0.59 \pm 0.08$ & $0.37 \pm 0.07$ \\
Co-Training (Offline RL) & $0.63 \pm 0.08$ & $0.17 \pm 0.14$ \\
\midrule
BIFROST                  & $0.68 \pm 0.08$ & $0.50 \pm 0.08$ \\
\bottomrule
\end{tabular}
\end{table}

As shown in Table~\ref{tab:sim2sim}, BIFROST achieves the highest success rate in both settings, though the margin varies. In the top-down experiment where the visual gap dominates, BIFROST ($0.68 \pm 0.08$) performs comparably to BDA ($0.67 \pm 0.05$), suggesting that a well-trained visual adaptor suffices when the dynamics gap is small. The advantage of bisimulation-based alignment becomes clear in the egocentric setting, where BIFROST ($0.50 \pm 0.08$) substantially outperforms BDA ($0.34 \pm 0.05$). Separately adapting visual and dynamics components does not capture the compound gap arising from simultaneous discrepancies between simulator backends.

Co-training with BC consistently improves over Target-Only in both settings, confirming that combining cross-domain expert data provides useful learning signal even without explicit alignment. Co-training with IQL further improves over BC in the top-down setting ($0.63$ vs.\ $0.59$), but becomes highly unstable in the egocentric setting ($0.17 \pm 0.14$), comparable to Target-Only ($0.16$). The large observation distribution mismatch between source and target causes the Q-function to propagate erroneous value estimates across domains during bootstrapping. The resulting instability negates the benefit of additional data. Direct Transfer yields near-zero success in the egocentric setting, confirming the severity of the domain gap.

\begin{figure}[t]
    \centering
    \includegraphics[width=\linewidth]{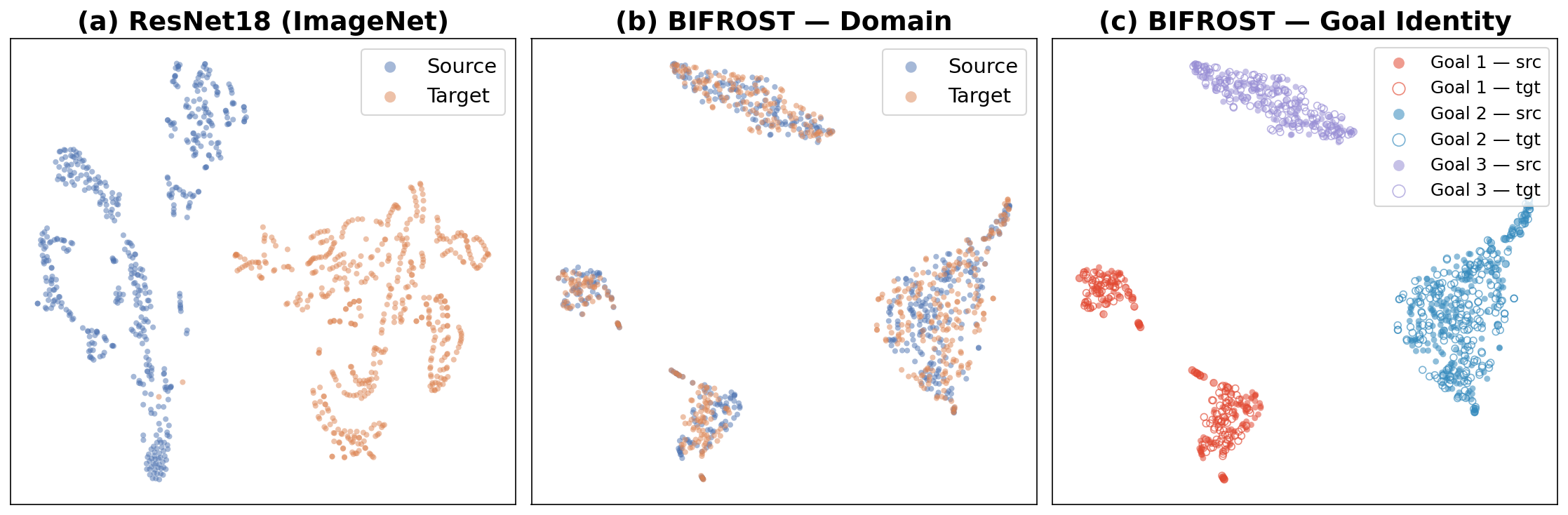}
    \caption{A t-SNE visualization on egocentric navigation. (a) ImageNet-pretrained ResNet18 features (512-dim) show domain-separated clusters. (b,c) BIFROST embeddings colored by domain and by goal identity, respectively. Goal 1 (red) forms two clusters corresponding to corridors separated by a maze wall.}
    \label{fig:tsne}
\end{figure}

Figure~\ref{fig:tsne} provides qualitative confirmation of the alignment mechanism. Even strong pretrained visual features fail to bridge the cross-simulator gap, producing fully domain-separated clusters. The BIFROST encoder collapses this separation into shared clusters that organize by goal identity rather than domain origin, confirming that the learned alignment preserves task-relevant structure.

\subsection{Sim2Real Transfer: Tabletop Manipulation}

Both sim2real experiments use a SO-101 low-cost manipulator arm. The source domain is built in SAPIEN; the target domain is the physical robot observed by an overhead $128 \times 128$ RGB camera. Target-domain data consists of 200 teleoperated trajectories, yielding approximately 6,000 paired segments of average length 45.

\subsubsection{Push T}

A T-shaped block must be pushed to a target pose on a tabletop. The agent observes the overhead RGB image along with the robot's joint angles, and commands a 2D end-effector position $(x, y)$ at fixed height, executed through an inverse kinematics solver. The domain gap arises from differences in friction, contact dynamics, and the actuation noise inherent to the low-cost arm.

\subsubsection{Visual Servoing}

The arm must move from an arbitrary configuration to a target configuration specified by a reference image. The observation is formed by concatenating the current and target overhead views; no proprioceptive input is provided, so the agent must infer the arm's state entirely from vision. The action is a joint position command. The reward measures the distance between current and target joint configurations, read from ground-truth state in simulation and from joint encoders on the physical robot. The dominant gap is in dynamics: the simulated actuator model does not capture the backlash, friction, and compliance of the physical servos.

\begin{table}[t]
\caption{Sim2Real manipulation results (mean $\pm$ std over 3 seeds).}
\label{tab:sim2real}
\centering
\setlength{\tabcolsep}{4pt}
\begin{tabular}{lcc}
\toprule
Method & Push-T & Vis. Servoing \\
       & Succ. \%$\uparrow$ & Joint Err. (deg)$\downarrow$ \\
\midrule
Direct Transfer          & $0.18 \pm 0.02$ & $10.41 \pm 1.78$ \\
Target-Only              & $0.25 \pm 0.08$ & $9.22 \pm 1.16$ \\
BDA                      & $0.42 \pm 0.09$ & $8.19 \pm 0.75$ \\
Co-Training (BC)         & $0.33 \pm 0.08$ & $9.16 \pm 1.24$ \\
Co-Training (Offline RL) & $0.28 \pm 0.12$ & $11.79 \pm 1.40$ \\
\midrule
BIFROST                  & $0.50 \pm 0.07$ & $7.64 \pm 0.33$ \\
\bottomrule
\end{tabular}
\end{table}

As shown in Table~\ref{tab:sim2real}, BIFROST achieves the best performance on both tasks with the lowest variance across seeds, outperforming BDA by 8 percentage points on Push-T and reducing joint error by 0.55 degrees on visual servoing. In the sim2sim experiments, the domain gaps are relatively controllable since both domains are simulators with known parameters; in this regime BDA performs comparably to BIFROST on top-down navigation. The sim2real setting introduces gaps that are harder to characterize and decompose, including unmodeled friction, contact compliance, and actuator backlash, widening the advantage of bisimulation alignment, which operates on behavioral consequences rather than requiring explicit modeling of individual gap sources. Co-Training (BC) offers only marginal improvement over Target-Only on both tasks, indicating that pooling cross-domain expert demonstrations without alignment provides limited benefit when the observation distributions diverge substantially. Co-Training (Offline RL) again degrades severely, producing higher error than Direct Transfer on visual servoing, confirming that Q-function bootstrapping across mismatched domains is harmful in both sim2sim and sim2real settings. The consistent improvement across Push-T, where the dominant gap involves contact dynamics, and visual servoing, where the gap is primarily in actuator modeling, suggests that the alignment mechanism generalizes across gap types.

\subsection{Ablation Study}
To evaluate the contribution of each algorithmic component in BIFROST, we conduct an extensive ablation study on egocentric visual navigation. We evaluate the impact of our loss formulations and the choice of the optimal transport metric.

\paragraph{Component-wise Contribution}
We compare the full BIFROST objective against two variants:
\begin{enumerate}
    \setlength{\itemsep}{0pt}
    \setlength{\topsep}{0pt}
    \item w/o Reward, which removes the shared target-reward prediction objective $\sL_{\text{reward}}$;
    \item w/o Latent Dynamics Prediction (LDP), which omits the self-predictive loss $\mathcal{L}_{\text{ldp
    }}$.
\end{enumerate}

\begin{table}[h]
\centering
\caption{Ablation results of different objective variants. Results are reported as mean (std) over 10 seeds.}
\label{tab:ablation_results}
\begin{tabular}{lcc}
\toprule
Variant & Source Return & Target Return (Zero-shot) \\
\midrule
BIFROST (full)                         & $0.4736\pm0.3382$ & $0.2511 \pm 0.0951$\\
W/o Reward ($\sL_{\text{reward}}$)     & $0.5144\pm0.2138$ & $0.2136 \pm 0.1420$\\
W/o LDP ($\sL_{\text{ldp}}$)           & $0.4532\pm0.3510$ & $0.2315 \pm 0.1204$\\
\bottomrule
\end{tabular}
\end{table}

Both variants retain $\sL_{\text{align}}$, as cross-domain alignment is the defining mechanism of BIFROST; its removal reduces the method to co-training, already evaluated in Table~\ref{tab:sim2sim}. As shown in Table~\ref{tab:ablation_results}, neither removal is catastrophic, but both degrade transfer. The w/o Reward variant exhibits a pattern consistent with the MSE metric ablation in Table~\ref{tab:metric_ablation}: highest source return ($0.5144$) paired with lowest target return ($0.2136$) and elevated cross-seed variance. Without target-reward supervision, the encoder gains additional freedom to specialize to source-domain features, improving source performance at the cost of transfer. The w/o LDP variant degrades more gracefully: the reward predictor still grounds the latent space in task-relevant outcomes, but the absence of dynamics prediction weakens the structure that $\sL_{\text{align}}$ operates over. The full objective achieves the highest target return with the lowest variance, indicating that the two losses play complementary roles: $\sL_{\text{reward}}$ anchors the latent space to target-domain task outcomes and mitigates source overfitting, while $\sL_{\text{ldp}}$ supplies the dynamical structure necessary for effective cross-domain alignment.

\paragraph{Choice of Optimal Transport Metric}
A key design choice in BIFROST is the use of the Wasserstein-1 distance ($W_1$) for distribution alignment. We compare our dual-form $W_1$ implementation against two alternatives in Table~\ref{tab:metric_ablation}:
(i) MSE-Alignment: utilizing the Mean Squared Error between the mean vectors $\mu(z^{\text{src}}, a)$ and $\mu(z^{\text{tgt}}, a)$; 
(ii) $W_2$-Closed-Form, using the closed-form Wasserstein-2 ($W_2$) distance between diagonal Gaussians:
\begin{equation}
    \begin{aligned}
    W_2^2\!\left(\hat{P}(\cdot \mid z, a),\; \hat{P}(\cdot \mid z', a)\right)
    &= \|\mu(z,a) - \mu(z',a)\|^2 \\
    &\quad + \|\sigma(z,a) - \sigma(z',a)\|^2,
    \end{aligned}
\end{equation}
which penalizes differences in both means and standard deviations.

\begin{table}[h]
\centering
\caption{Ablation results of different metrics for $\sL_{\text{align}}$. Results are reported as mean (std) over 10 seeds.}
\label{tab:metric_ablation}
\begin{tabular}{lcc}
\toprule
Metric & Source Return & Target Return (Zero-shot) \\
\midrule
BIFROST ($W_1$ Dual) & $0.4736\pm0.3382$ & $0.2511 \pm 0.0951$ \\
$W_2$ (Closed-form)  & $0.4424\pm0.3729$ & $0.2058 \pm 0.1362$ \\
MSE (Mean only)      & $0.5133\pm0.2179$ & $0.1263 \pm 0.1099$ \\
\bottomrule
\end{tabular}
\end{table}

The three metrics form a clear hierarchy in transfer performance. MSE alignment achieves the highest source return yet the lowest target return, a signature of source-domain overfitting. Because MSE leaves the variance parameters $\sigma(z,a)$ entirely unconstrained, the forward model can encode domain identity into the variance structure while satisfying the alignment loss, resulting in a latent space that appears aligned in its first moment but remains separated in its distributional structure. The $W_2$ closed-form metric corrects this by additionally penalizing $\|\sigma_{\text{src}} - \sigma_{\text{tgt}}\|^2$, improving target return from $0.1263$ to $0.2058$. The $W_1$ dual formulation yields a further improvement to $0.2511$ with lower cross-seed variance, as its gradient magnitude does not diminish when the two distributions converge, providing a more sustained alignment signal during late-stage training.

\section{Limitations And Discussions}
\label{sec:limitations}

BIFROST's effectiveness rests on the quality of the paired data constructed in Section~\ref{sec:paired}, and the current construction procedure introduces assumptions that constrain the scope of applicability.
The first concerns dynamics continuity. In tasks involving contact discontinuities such as grasping or insertion, millimeter-scale differences in initial pose or contact stiffness can produce binary outcome divergence: the physical gripper closes successfully while the simulated gripper slips, or vice versa. Beyond the bifurcation point, the two trajectory segments are no longer behaviorally equivalent, yet the alignment loss $\mathcal{L}_{\text{align}}$ continues to enforce proximity between their temporally paired latent states. The resulting training signal is actively harmful, as the encoder is supervised to map divergent outcomes to adjacent latent representations. 
% Addressing this limitation requires moving beyond strict temporal pairing toward correspondence based on behavioral similarity, or introducing closed-loop replay in simulation to periodically correct the simulator state using real observations.
%
The second concerns the implicit dependence on state estimation during data collection. Although BIFROST operates strictly from raw observations at inference time, constructing paired data requires resetting the simulator to the observable configuration $\xi_i$ extracted via AprilTags. This dependence is weaker than requiring full ground-truth state, but the pipeline is not fully end-to-end during data collection.
Finally, because the shared history encoder is trained on offline paired data, its representational coverage is bounded by $\mathcal{D}_{\text{paired}}$. Online exploration during policy learning may reach states absent from the training data, where the encoder produces unreliable latent representations that the policy can exploit.
These limitations originate primarily from the data construction pipeline. A natural next step is to replace strict temporal pairing with trajectory-level alignment via optimal transport, which would accommodate dynamics discontinuities by matching trajectory segments based on behavioral similarity rather than timestep correspondence.

% Something to be added

\section*{ACKNOWLEDGMENT}
This work took place in the Prediction and Action Lab (PAL) at the University of Wisconsin--Madison. PAL research is supported by NSF (IIS-2410981) and the Wisconsin Alumni Research Foundation. 
The authors thank Brahma S. Pavse and Sangwoo Shin for helpful discussions and feedback on the manuscript.

%%%%%%%%%%%%%%%%%%%%%%%%%%%%%%%%%%%%%%%%%%%%%%%%%%%%%%%%%%%%%%%%%%%%%%%%%%%%%%%%

% References are important to the reader; therefore, each citation must be complete and correct. If at all possible, references should be commonly available publications.

\bibliographystyle{IEEEtran}
\bibliography{ref}

\end{document}